%% file: ms.tex
\documentclass[10pt,conference,letterpaper]{IEEEtran}
\IEEEoverridecommandlockouts

\usepackage[english]{babel}
\usepackage[utf8]{inputenc}
\usepackage[T1]{fontenc}

\usepackage{amssymb}
\usepackage{amsmath}
\usepackage{graphicx}
\graphicspath{{figures/}{plots/}}
\usepackage[scaled]{beramono}
\usepackage[numbers]{natbib}

\usepackage{array}
\usepackage{subfig}
\usepackage{datetime}
\usepackage{caption}
\usepackage{etoolbox}
\usepackage[detect-all,binary-units]{siunitx}
\usepackage{blkarray}
\usepackage{multirow}
\usepackage{makecell}

\usepackage{tabularray}
\UseTblrLibrary{booktabs} 

\usepackage{acro}
\DeclareAcronym{HPC}{short=HPC, long=high performance computing}
\DeclareAcronym{NN}{short=NN, long=neural network}
\DeclareAcronym{ML}{short=ML, long=machine learning}
\DeclareAcronym{DSP}{short=DSP, long=digital signal processing}
\DeclareAcronym{UAM}{short=UAM, long=urban air mobility}
\DeclareAcronym{CNN}{short=CNN, long=convolutional neural network}
\DeclareAcronym{FPGA}{short=FPGA, long=field-programmable gate array}
\DeclareAcronym{ONNX}{short=ONNX, long=Open Neural Network Exchange}
\DeclareAcronym{FAA}{short=FAA, long=Federal Aviation Administration}
\DeclareAcronym{EASA}{short=EASA, long=European Union Aviation Safety Agency}
\DeclareAcronym{CPU}{short=CPU, long=central processing unit}
\DeclareAcronym{GPU}{short=GPU, long=graphical processing unit}
\DeclareAcronym{TPU}{short=TPU, long=tensor processing unit}
\DeclareAcronym{RNE}{short=RNE, long=round to nearest even}
\DeclareAcronym{RNA}{short=RNA, long=round to nearest away}
\DeclareAcronym{RTZ}{short=RTZ, long=round towards zero}
\DeclareAcronym{MAC}{short=MAC, long=multiply–accumulate}
\DeclareAcronym{OPS}{short=OPS, long=operations per second}
\DeclareAcronym{FLOPS}{short=FLOPS, long=floating-point operations per second}
\DeclareAcronym{LUT}{short=LUT, long=lookup table}
\DeclareAcronym{FF}{short=FF, long=flip-flop}

\author{
	\IEEEauthorblockN{Fabien Geyer}
	\IEEEauthorblockA{
		Airbus Central R\&T \\
		Munich, Germany}
	\and
	\IEEEauthorblockN{Johannes Freitag}
	\IEEEauthorblockA{
		Airbus Central R\&T \\
		Munich, Germany}
	\and
	\IEEEauthorblockN{Tobias Schulz}
	\IEEEauthorblockA{
		Airbus Central R\&T \\
		Munich, Germany}
	\and
	\IEEEauthorblockN{Sascha Uhrig}
	\IEEEauthorblockA{
		Airbus Central R\&T \\
		Munich, Germany}
}

\usepackage{flushend}
\usepackage{nameref}
\usepackage[capitalize,noabbrev]{cleveref}

\begin{document}

\title{Efficient and Mathematically Robust Operations for Certified Neural Networks Inference}

\maketitle
\acresetall

\input{content}

{
\footnotesize
\yyyymmdddate
\bibliographystyle{IEEEtran}
\bibliography{IEEEabrv,biblio}
}

\end{document}

%% file: content.tex
\begin{abstract}
In recent years, \ac{ML} and \acp{NN} have gained widespread use and attention across various domains, particularly in transportation for achieving autonomy, including the emergence of flying taxis for \ac{UAM}.
However, concerns about certification have come up, compelling the development of standardized processes encompassing the entire \ac{ML} and  \ac{NN} pipeline.
This paper delves into the inference stage and the requisite hardware, highlighting the challenges associated with IEEE 754 floating-point arithmetic and proposing alternative number representations.
By evaluating diverse summation and dot product algorithms, we aim to mitigate issues related to non-associativity.
Additionally, our exploration of fixed-point arithmetic reveals its advantages over floating-point methods, demonstrating significant hardware efficiencies.
Employing an empirical approach, we ascertain the optimal bit-width necessary to attain an acceptable level of accuracy, considering the inherent complexity of bit-width optimization.
\end{abstract}
\acresetall

\section{Introduction}

The last few years have seen an increasing use of \ac{ML} and \acp{NN} in many domains, mainly due to their good results on challenges where no other solutions has yet been found.
One important domain is transportation, where \acp{NN} are seen as the way forward for bringing autonomy, not only for cars, but also for other modes of transportation such as flying taxis for \ac{UAM} or more generally aviation.

Various concerns about certification of \ac{ML} and \acp{NN} have been raised by the aeronautical regulation and certification bodies \cite{AIR6988,ER022,EASARoadmap,NASARoadmap}, especially for functions seen as safety critical.
Similar concerns have also been raised in other transportation industries as shown by a recent survey \cite{PerezCerrolaza2023}.
To overcome these concerns, methods for certification of the complete \ac{ML} and \acp{NN} pipeline are currently under development~\cite{CoDANN2020,CoDANN2021} and will be standardized in the future~\cite{ARP6983}.
The certification process covers the aspects of a \ac{ML} pipeline, including the software aspects such as data verification and validation, training, inference, and tests. Of course, the embedded hardware used for inference must show appropriate characteristics such as numerical accuracy and performance, while being compliant with the related hardware standards.

In this paper, we focus on the inference part and the hardware used for it.
This is a challenging task as the hardware used for training -- mainly clusters of \acp{GPU} or \acp{TPU} -- is generally vastly different from the one used for inference.
The training and inference hardware are usually at different optimum points when considering the different trade-offs that have to be made in terms of efficiency, flexibility, power, memory footprint, environmental conditions, and predictability.
Additionally for the aeronautical industry, certification of the inference hardware is required, meaning that predictable execution time and mathematical robustness is mandatory.
In practice, this means that a good trade-off between performance (e.g., inference rate), power consumption, complexity, and numerical accuracy must be found in addition to predictable execution time.

To meet these challenges, custom hardware based on \acp{FPGA} is a promising solution since it provides the flexibility to manage these trade-offs, including the execution predictability.
Moreover, an \ac{FPGA} design comes as a white box being a clear advantage for certification.
We investigate in this paper some key aspects about mathematical robustness when using \acp{FPGA} for accelerating \acp{NN}, namely: the challenges of using IEEE 754 floating-point arithmetic and alternate number representation as a solution to those.

The main challenge of IEEE 754 floating-point arithmetic is that it is not associative, meaning that the order of operations can influence the final result of the computation.
In an aeronautical certification context, this can lead to a challenge as hardware with a significantly higher mathematical performance is used for training compared to the inference device embedded into an airborne vehicle.

In this paper, we will investigate some methods to alleviate this issue by evaluating different summation and dot product algorithms for floating-point arithmetic.
Moreover, we will investigate the use of fixed-point arithmetic, as this alternative number representation format does not suffer from the issue with the associativity of the operations.
We will illustrate that it also enables non negligible gains in terms of hardware requirements, as operations on integers are simpler to implement than IEEE 754 floating-point arithmetic.
We will also investigate how many bits are required to reach a sufficient accuracy by an empirical approach, as bit-width optimization is known to be NP-hard \cite{Constantinides2002}.

Finally, we will evaluate the impact of floating point and fixed point arithmetics in terms of hardware resources for two exemplary \acp{CNN} for image classification.
Our goal is to find the optimum point in terms of numerical accuracy, bit-width, and hardware resources for an exemplary \ac{FPGA} platform.

This paper is organized as follows.
First, we review the related work in \cref{sec:relatedwork}.
\cref{sec:certification} provides some background on the challenges of certification of \ac{ML} in the aeronautical industry.
We review existing methods for accurate floating-point in \cref{sec:floatingpoint} and fixed-point computations in \cref{sec:fixedpoint}, and the hardware resources needed for them in \cref{sec:hardware:resources}.
A numerical evaluation of our approach is shown in \cref{sec:evaluation}.
Finally, \cref{sec:conclusion} concludes the paper.

\section{Related work}
\label{sec:relatedwork}

A recent survey on \acp{NN} approximations for custom hardware \cite{Wang2019} shows that different methods have been proposed in the literature to speed-up the inference of \acp{NN}, such as alternative number representation, quantization, pruning or activation function approximation.
Similary, \cite{Cherubin2020} recently surveyed tools for reduced precision computation, a growing trend for enhancing performance metrics in embedded systems and \ac{HPC}. It highlights the lack of automated precision customization support in standard compiler frameworks and the ongoing research to improve automation, emphasizing the need for better tools, especially those based on static analysis.

Finding the appropriate number representation for \acp{NN} has already been extensively investigated in the literature, with the evaluation of bfloat16 \cite{Kalamkar2019,Burgess2019}, posits \cite{Carmichael2019,Cococcioni2021,Lu2021}, 
Microsoft floating point \cite{Rouhani2020},
FlexPoint \cite{Koster2017},
or adaptive floating-points \cite{Liu2021}.

In the scope of fixed-point operations, various works focused on integer-only inference and training for \acp{NN}.
FxpNet is proposed in~\cite{Chen2017}, a framework for training deep \acp{CNN} using low bit-width arithmetics in both forward and backward passes, adapting the bit-width of stored parameters during training.
It employs integer batch normalization and fixed-point optimization methods to minimize floating-point operations, leading to power and chip area savings, with experimental results demonstrating comparable accuracy to state-of-the-art binarized and quantized approaches.
\cite{Zhen2019,Yao2021}~recently introduced HAWQ and HAWQV3, an integer-only inference where the entire computational graph is performed only with integer operations.
They address the hidden costs of current low-precision quantization algorithms, and present a novel mixed-precision integer-only quantization framework that enables integer-based computations and hardware-aware quantization.

Finding the optimal bitwidth given a mathematical formula has been a challenge since the age of \acp{DSP}.
\cite{Lee2006} proposed an automated static method for optimizing bit widths of fixed-point feedforward designs, ensuring guaranteed accuracy.
It employs semi-analytical precision analysis and adaptive simulated annealing to minimize both integer and fraction parts.
In the scope of \acp{NN}, \cite{Ioualalen2019}~proposed a novel technique involving linear programming and integer variables to optimize \acp{NN} precision without compromising output quality beyond a user-defined threshold.
It is based on the method from \cite{Martel2017}, which combines forward and backward static analyses through abstract interpretation, expressed as a set of constraints with first-order predicates and affine integer relations, simplifying verification by an SMT Solver.
A similar approach was applied to code generation for \acp{NN} using error analysis in~\cite{Benmaghnia2022,Benmaghnia2022b}.
While the methods from~\cite{Ioualalen2019,Benmaghnia2022,Benmaghnia2022b} were shown to be successful on small \acp{NN}, they scale poorly to larger networks like ResNet, as shown later in \cref{sec:mathematical_bound}.

\section{Background on certification}
\label{sec:certification}

Certification of hardware and software in the aeronautical industry is a rigorous process aiming at ensuring the safety, reliability, and compliance of aviation systems with stringent regulatory standards.
It involves thorough testing, analysis, and documentation to verify that the onboard equipment and software meet strict requirements set by aviation authorities such as the \ac{EASA} or the \ac{FAA}.
This certification process plays a critical role in guaranteeing the airworthiness of aircraft, promoting technological advancements, and maintaining the highest levels of safety for passengers and crew.

In addition to hardware and software certification, the aeronautical industry is increasingly focusing on the incorporation of onboard \ac{ML} and its robustness in critical aviation systems.
Ensuring the reliability and efficiency of \ac{ML} algorithms and their hardware implementation is crucial for tasks such as predictive maintenance, autonomous decision-making, and enhanced flight operations.
This necessitates a comprehensive evaluation of the \ac{ML} models' performance under various operational scenarios, as well as a meticulous examination of the hardware used for inference.

\begin{figure}[h!]
	\includegraphics[width=\columnwidth]{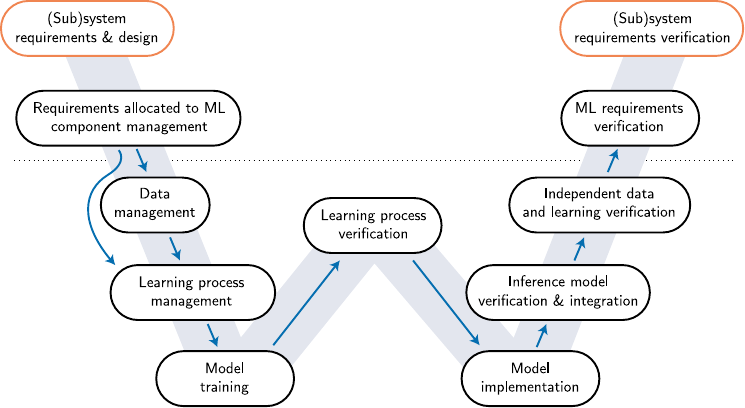}
	\caption{W-shaped development cycle for design assurance for \acp{NN} from \cite{CoDANN2021}}
	\label{fig:wshaped-development}
\end{figure}

To address this challenge, a W-shaped development process as illustrated in \cref{fig:wshaped-development} has been proposed for tailor the classical V-shaped cycle to \ac{ML} applications \cite{CoDANN2020,CoDANN2021}.
Various efforts have been started to standardize this development process in order to meet the high reliability and robustness requirements \cite{AIR6988,ER022,EASARoadmap,NASARoadmap}.

In this paper we focus on the challenges for the development of hardware accelerators used for onboard \acp{NN} inference.
These accelerators must exhibit sufficient performance to enable execution of large \acp{NN} at high enough execution rate by employing techniques such as pipelining, parallelization and numerical approximations.
However, the necessity for heightened efficiency clashes with the demand for precise numerical accuracy caused by limited computational hardware resources.

\section{Floating-point arithmetic}
\label{sec:floatingpoint}

Floating-point arithmetic is a method of representing real numbers in a way that allows a wide range of values to be expressed using a fixed number of bits.
It has been standardized under IEEE 754 \cite{IEEE754}, a widely accepted standard that defines the format and rules for performing arithmetic operations with floating-point numbers.
It is commonly found in off-the-shelves \acp{CPU}, \acp{GPU} and \acp{TPU}.

Despite this standardization, issues related to the order of operations and precision limitations persist.
Due to the finite precision of floating-point numbers, operations like addition, subtraction, multiplication, and division may not always yield exact results, leading to rounding errors and loss of precision, especially when performed in a different order than intended.
This can potentially impact the accuracy of numerical computations, making it crucial to be aware of these limitations when working with floating-point numbers.

As parallelism is often used for accelerating computations, the order of operations is not guaranteed, leading to numerical inaccuracies.
This holds significance not just within the realm of certification of \acp{NN} but also in other domains like scientific computing, where the reproducibility of results is considered crucial \cite{Apostal2020,Robey2011}.
In addition to several rounding methods, we evaluate different implementations of operations commonly found in \acp{NN}, namely summation and dot products.

\subsection{Rounding}
\label{sec:rounding}

IEEE 754 defines several rounding modes, which determine how a floating-point number should be rounded to fit into a specific precision.
The default rounding mode is \emph{\ac{RNE}}, where numbers are rounded to the nearest representable value. If the number falls exactly midway between two representable values, it chooses the one with an even least significant bit.

In this paper, we will also evaluate two additional rounding modes:
\emph{\ac{RTZ}} which always truncates the fractional part, effectively rounding towards zero regardless of the sign of the number;
and \emph{\ac{RNA}}, where numbers are rounded to the nearest representable value, and if it is equidistant from both, it is rounded away from zero.

\subsection{Summation algorithms}

The order of operations is crucial when summing floating points to ensure accurate results and prevent rounding errors that could accumulate with each operation.
Various methods have been proposed to address this, as showed in \cite{Higham1993}.
For this paper, we focus on four approaches, as they are easily implemented in hardware and do not require any sorting.

The \emph{naive accumulation} summation algorithm, also known as the straightforward or simple summation method, involves iteratively adding each element of a given set of numbers to an accumulator or running total. 

The \emph{pairwise summation} algorithm recursively divides the set of numbers into pairs, adding the pairs together, and then continuing the process until a single sum is obtained.

The \emph{Kahan summation} algorithm \cite{Kahan1965} and its extensions \cite{Neumaier1974}, also known as compensated summation algorithms, keep track of the accumulated error during the summation process and compensating for it during each step of the process.

Finally, the \emph{exact summation} algorithm corresponds to a fixed-point accumulator used for summing floating point introduced by \cite{Dinechin2008}.
This method is also detailed later in \cref{sec:hardware:resources}.

\subsection{Dot product algorithms}

Similar to the compensated summation algorithm, a compensated dot product algorithm -- labeled \emph{ORO} dot product in the text -- has been proposed in \cite{Ogita2005}.
The algorithm uses error-free transformations of the sum and product of two floating point numbers to perform accurate dot products.

\section{Fixed-point arithmetic}
\label{sec:fixedpoint}

Quantizing the operations to 8 bits integers \cite{Jacob2018} is currently gaining more traction due to its efficiency on CPUs and hardware accelerators.
Yet, this approach isn't without drawbacks: it requires a post-processing phase of the trained \ac{NN} to scale the numbers, and the drop in performance can be significant in some cases as shown later in \cref{sec:evaluation:int8}.

We investigate a similar approach using fixed-point arithmetic, a method widely used in \ac{DSP} and gaming due to their speed compared to floating points.
Our approach does not require scaling of the weights of the \ac{NN} and is simpler from a computational of view than using floating points.
Additionally, the order of the operations is not relevant here, compared to floating point.

\subsection{Definition}

Fixed-point arithmetic is a method of representing numbers by storing a fixed number of digits of their fractional part.
Numbers are represented as integers which are split into three parts: a sign bit, a magnitude part with $m$ bits and a fractional part with $f$ bits.
Conversion from a real number $x$ to its fixed point representation is done via the following function:
\begin{equation} \label{eq:float2fixpoint}
	\mathit{round}\left(x \cdot 2^f\right)
\end{equation}
with $\mathit{round}$ a rounding function as described in \cref{sec:rounding}.

Mathematical operations can be easily performed using the underlying integers.
To add or subtract two values of the same fixed-point type, it is sufficient to add or subtract the underlying integers.

To multiply two fixed-point numbers, it suffices to multiply the two underlying integers, giving a result with a fractional part of $2f$ bits.
To avoid an increasing number of bits for the fractional part when performing multiple multiplications, rescaling is required.
This is performed by shifting right the underlying integer and taking care of rounding.

As illustrated later in \cref{sec:evaluation:fixp,fig:resnet18-imagenet-fixp-dotproductfunction-top1same}, correct rounding for the multiplication operation can dramatically improve the results.

\subsection{Dot product algorithm}

From the description of the previous section, the main loss of information appears during the multiplication of two fixed-point numbers, where a right shift and rounding operation is required.
We reformulate the dot product of the $\mathbf{x}$ and $\mathbf{y}$ vectors as:
\begin{align}
	\mathbf{x} \cdot \mathbf{y} & = \sum_i \mathit{SR}(x_i y_i)            & \text{\it (naive dot product)}  \label{eq:fixp-naive-dotproduct} \\
	                            & = \mathit{SR}\left(\sum_i x_i y_i\right) & \text{\it (accurate dot product)} \label{eq:fixp-accurate-dotproduct}
\end{align}
with $x_i$ and $y_i$ the underlying integer at position $i$ of the vectors $\mathbf{x}$ and $\mathbf{y}$, and $\mathit{SR}$ the right shift and round operation.

As only one rounding operation is required in \cref{eq:fixp-accurate-dotproduct}, it will produce more accurate results than \cref{eq:fixp-naive-dotproduct}.

\subsection{Mathematically bounding the error}
\label{sec:mathematical_bound}

As mentioned in \cref{sec:relatedwork}, the works from \cite{Benmaghnia2022,Benmaghnia2022b} propose a method based on affine arithmetic in order to mathematically bound the absolute error of the outputs of an \ac{NN}.
It can easily be derived from \cref{eq:float2fixpoint} that the magnitude of the maximum difference between a real value and it's representation is $2^{-f}$.
Propagating this error through the operation of the \ac{NN} is then performed using affine arithmetic.

\begin{figure}[h!]
	\includegraphics[width=\columnwidth]{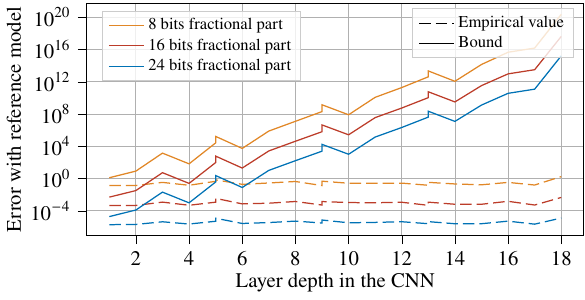}
	\caption{Evaluation of mathematical bounds and empirical values of the error of executing and converting the ResNet18 \ac{CNN} \cite{He2016} in fixed point arithmetic}
	\label{fig:resnet18-testimage-fixp-bound}
\end{figure}

While such method is applicable on relatively small \acp{NN} with only a few layers, it becomes impossible to use on deeper architectures.
To illustrate this, we computed the bounds given by this method on a pretrained ResNet18 \ac{CNN} \cite{He2016} using fixed point arithmetics with different bit widths.
The results are shown in \cref{fig:resnet18-testimage-fixp-bound}, where the error propagation of the \ac{CNN} trends to get larger with increasing number of layers. Accordingly, a static analysis is not useful in practice and not further evaluated in this paper.

\section{Resource usage for arithmetic}
\label{sec:hardware:resources}

We evaluate here the resources required for different binary number formats in \acp{FPGA}.
Due to their flexibility, \acp{FPGA} offer a wider range of formats than the commonly found ones in off-the-shelve \acp{CPU} or \acp{GPU} such as 8, 16, 32 or 64 bits integers or floating-points.
Furthermore, the formats can be adjusted for the specific \acp{NN} that shall be executed depending on the precision that is needed for the given application.
In order to perform the multiplication and adding operations needed, essential parts of any \acp{NN} processor are the \ac{MAC} units. 
These hardware blocks perform the multiplication of two values and accumulate the result of the multiplication.
The number of \ac{MAC} units which can be utilized in parallel at a certain clock frequency determines the maximum achievable performance of the device.
The performance is typically given in \ac{OPS} or \ac{FLOPS}. As one \ac{MAC} operation consists of a multiplication and an accumulation, both computations are counted separately.

\Acp{FPGA} offer different resources that can be configured and connected to implement the intended logic. 
The main resources, available in all different types of \acp{FPGA}, are \acp{LUT} which are configured to implement the combinatorial logic and \acp{FF}.
In addition, \acp{DSP} can be integrated in the design to speed up specific operations commonly needed for example in filters, fast-Fourier-transforms or other suitable algorithms. 
These \acp{DSP} differ depending on the \ac{FPGA} architecture and vendor. 
\Acp{FPGA} are available in different sizes, different numbers of \acp{LUT}, \acp{FF},  \acp{DSP} and the relation between the elements needed can be selected. 
There are \acp{FPGA} with a high number of \acp{LUT} compared to available \acp{DSP} and vice versa in order to select the right \ac{FPGA} for the task as certain designs might be able to leverage \acp{DSP} while this is not possible in a different design. 
For our evaluations we selected the AMD VU9P, based on the Virtex Ultrascale Plus technology, because of the balanced relation between \acp{LUT}, \acp{FF} and \acp{DSP} \cite{AMD2023}.

\begin{figure}[h!]
	\includegraphics[width=\columnwidth]{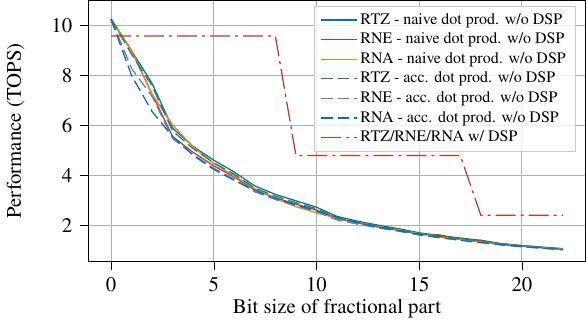}
	\caption{Performance for fixed-point MACs on AMD VU9P FPGA with and without DSP usage}
	\label{fig:hw-int-performance}
\end{figure}

In this paper, we analyze the performance achievable on \acp{FPGA} for fixed-point as well as floating-point numbers of arbitrary size of the fractional part / mantissa.
For fixed-point, a \ac{MAC} unit consisting of the integer Multiplier and Adder/Subtractor AMD IP cores of Vivado 2023.1 was synthesized and implemented for different numbers of fractional bits while using 10 bits for the integer part, which has shown to be the necessary bit width to achieve the desired accuracy on ResNet18 without the need to scale the numbers. 
Additionally, variants were implemented for different rounding modes and whether the rounding is done after the accumulation (accurate dot product) or after every multiplication (naive dot product). 
The results of this analysis are the resource usages (number of \acp{LUT}, \acp{FF},  \acp{DSP}) and the maximum frequencies for a single \ac{MAC} unit on the aforementioned \ac{FPGA}. 
Furthermore, the results include resource usage for fabric only (without  \acp{DSP}) for a better transferability to other \ac{FPGA} architectures, and with \ac{DSP} usage to achieve the maximum performance on the given \ac{FPGA}. 
In order to transfer the resource usage and frequencies to performance estimates, it was theoretically analyzed how many of these \ac{MAC} units fit into the exemplary \ac{FPGA}.
It was assumed that \SI{70}{\percent} of the \acp{LUT} and \acp{FF} can be utilized to avoid potential timing closure issues and routing congestion while \SI{100}{\percent} of the \acp{DSP} can be used. 
The number of \ac{MAC} units fitting in the \ac{FPGA} is multiplied by the achievable frequency which defines the upper bound of the performance of the complete chip and is shown in \cref{fig:hw-int-performance}.

In the figure it is visible that in the case of no \acp{DSP} usage, for an increased number of bits the performance exponentially decreases while there is only a slight difference when comparing the different rounding and accumulation styles. 
However, with the usage of \acp{DSP}, the performance decreases in steps. 
In this case the \acp{DSP} are the limiting factor for the full utilization of the \ac{FPGA} and the number of \acp{DSP} used by the \ac{MAC} units of different bit sizes is increasing stepwise i.e. for fractional bit sizes of 9 to 17 the same number of \acp{DSP} per \ac{MAC} unit is needed.
Furthermore, there is no difference for the different rounding styles because the rounding only adds \acp{LUT} which are still available on the \ac{FPGA}. 
Thus, even though the amount of resources consumed are slightly different the resulting performance is identical.	

\begin{figure}[h!]
	\includegraphics[width=\columnwidth]{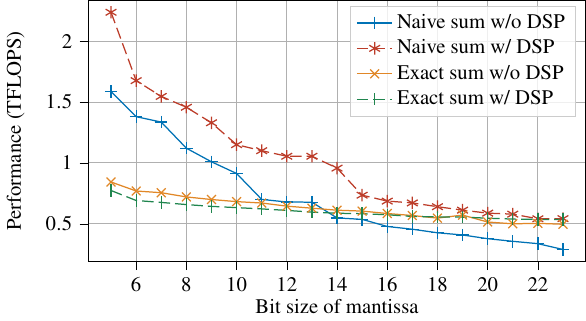}
	\caption{Performance for floating-point \acp{MAC} on AMD VU9P \ac{FPGA} with and without \ac{DSP} usage}
	\label{fig:hw-fp-performance}
\end{figure}

For floating-point, a \ac{MAC} unit was designed with the AMD floating-point IP core consisting of a multiplier and an adder implemented and synthesized for different bit sizes of the mantissa. 
However, using a floating-point adder as an accumulator, later referred to as \emph{naive} method, leads to a low frequency of the \ac{FPGA} as the addition has to happen within one cycle. 
A second \ac{MAC} unit was designed with the Floating-point IP core accumulator instead of the adder, later referred to as \emph{exact} method. 
This accumulator is implemented as a fixed-point accumulator internally which leads to a very high precision but also a very high resource consumption~\cite{Dinechin2008}.
However, since it can be pipelined, a high frequency is achievable. 
The theoretically achievable maximum performance on the given \ac{FPGA} are shown in \cref{fig:hw-fp-performance}.
As expected, the performance decreases for a higher number of bits for the mantissa. 
Contrary to the fixed-point analysis, if \acp{DSP} are used the bit size impacts the performance not stepwise because the \acp{LUT} are the dominant resource for the floating-point \ac{MAC} units.

\section{Numerical evaluation}
\label{sec:evaluation}

We numerically evaluate in this section the different approaches presented earlier and assess their impacts in terms of numerical accuracy and hardware resources.

\subsection{Methodology}

To evaluate the impact of the bitwidth and the different summation and dot product algorithms previously listed, we implemented our own \ac{ONNX} runtime using Go and C.
Floating point arithmetic are implemented using Go's \texttt{math/big.Float} arbitrary-precision arithmetic library.
We use our own implementation for fixed point arithmetic.
This enables us to precisely target the mathematical operations and number representation under investigation while being compatible with existing \ac{ONNX} toolchains.

As exemplary models for our numerical evaluation, we use the pretrained models from the \ac{ONNX} model zoo\footnote{\texttt{https://github.com/onnx/models}} for MNIST~\cite{LeCun1998} and ResNet18~\cite{He2016}.
For the numerical evaluations, the full test set of MNIST is used for the MNIST model, and a subset of ImageNet-1k dataset \cite{Deng2009} is used for ResNet18.

To evaluate the accuracy of our computations, we use the \ac{ONNX} runtime from Microsoft\footnote{\texttt{https://github.com/microsoft/onnxruntime}} with its \ac{CPU} execution provider as reference.
Our main metric is to evaluate if the top-1 classification from our model with lower precision is the same as the top-1 classification from the reference (32 bits floating-point) model.
For our evaluation and use-case, we aim at achieving a metric of \SI{100}{\percent}, i.e. match the classification from the reference.

\subsection{Evaluation of int8 quantized models}
\label{sec:evaluation:int8}

As a first benchmark, we evaluate the performance of int8 quantized models.
For the MNIST model, we use the already publicly available quantized version of the model from the \ac{ONNX} model zoo.
For ResNet18, we use the Intel Neural Compressor open-source tool \cite{IntelNeuralCompressor} to quantize the model.

\begin{table}[h!]
	\centering
	\caption{Metric for the int8 quantized models}
	\label{tab:results_int8}
	\begin{tabular}{lr}
		\toprule
		\textbf{Model} & \textbf{Same top-1 as reference} \\ \midrule
		ResNet18       &             \SI{85.80}{\percent} \\
		MNIST          &             \SI{54.69}{\percent} \\ \bottomrule
	\end{tabular}
\end{table}

Results are presented in \cref{tab:results_int8}.
It is clear from the values of our evaluation metric that the int8 quantized models are not sufficient since the required \SI{100}{\percent} metric is not reached.
These results justify why better precision and an evaluation of alternate approaches for accelerating inference are required for our use-cases.

\subsection{Floating-point arithmetic}
\label{sec:evaluation:bigf}

We evaluate in this section the performance with floating point arithmetic and the methods described in \cref{sec:floatingpoint}.

\subsubsection{Impact of the summation function}

\begin{figure}[h!]
	\includegraphics[width=\columnwidth]{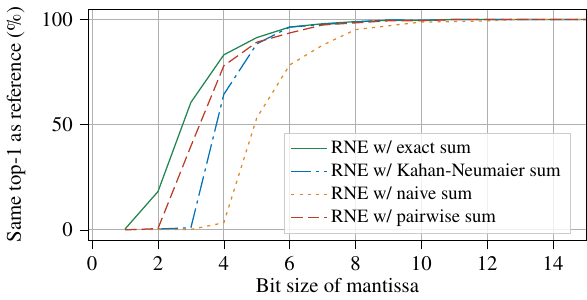}
	\caption{Impact of the summation algorithm on ResNet18 with floating-point arithmetic}
	\label{fig:resnet18-imagenet-bigf-sumfunction-top1same}
\end{figure}

Results illustrating the impact of the summation function (with naive dot product) on ResNet18 are presented in \cref{fig:resnet18-imagenet-bigf-sumfunction-top1same}.
There is a clear benefit at using more accurate summation functions, as it dramatically improves the accuracy of the results over the naive summation for low bit widths.

Overall, the exact summation provides the best results.
Yet, \SI{11}{\bit} for the mantissa are required for all three non-naive summation functions to achieve a metric of \SI{100}{\percent}.

\subsubsection{Impact of the dot product function}

\begin{figure}[h!]
	\includegraphics[width=\columnwidth]{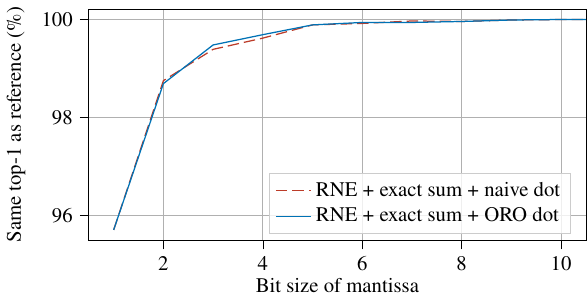}
	\caption{Impact of the dot product algorithm on MNIST with floating-point arithmetic}
	\label{fig:cntkmnist-mnist-bigf-dotproductfunction-top1same}
\end{figure}

Results illustrating the impact of the dot product function on MNIST are presented in  \cref{fig:cntkmnist-mnist-bigf-dotproductfunction-top1same}.
While the ORO dot product \cite{Ogita2005} enables us to gain a few percent on our metric, its impact is minimal: for both dot products, the same number of bits are required to reach \SI{100}{\percent}.
For ResNet18, the same conclusion can be made.

Overall, these results illustrate that adding the overhead of this dot product function in hardware is not worth it, as there is no gain in terms of bits required for computations.

\subsection{Fixed-point arithmetic}
\label{sec:evaluation:fixp}

We evaluate in this section the performance with fixed point arithmetic and the methods described in \cref{sec:fixedpoint}.

\subsubsection{Impact of rounding}

\begin{figure}[h!]
	\includegraphics[width=\columnwidth]{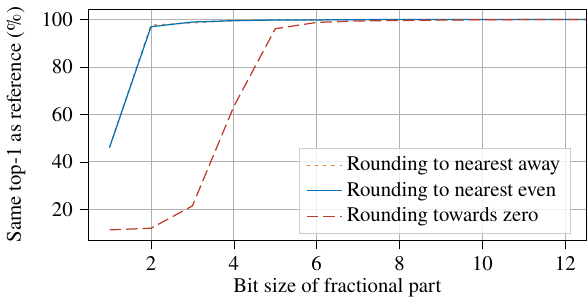}
	\caption{Impact of rounding mode on MNIST with fixed-point arithmetic with naive dot product. The \ac{RNE} and \ac{RNA} curves overlap.}
	\label{fig:cntkmnist-mnist-fixp-rounding-top1same}
\end{figure}

The impact of the rounding for the MNIST model is presented in \cref{fig:cntkmnist-mnist-fixp-rounding-top1same}.
Correctly rounding with \ac{RNE} or \ac{RNA} during the multiplication dramatically improves the results for low bit widths compared to \ac{RTZ}, saving \SI{3}{\bit} on average.

\subsubsection{Impact of the dot product function}

\begin{figure}[h!]
	\includegraphics[width=\columnwidth]{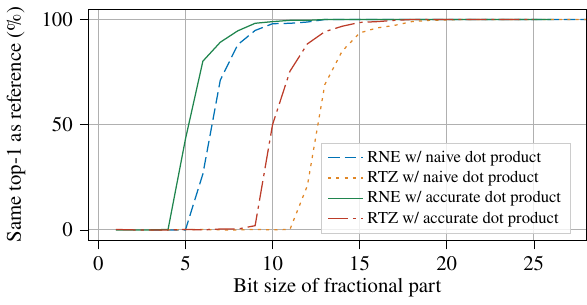}
	\caption{Impact of the rounding mode and dot product functions on ResNet18 with fixed-point arithmetic}
	\label{fig:resnet18-imagenet-fixp-dotproductfunction-top1same}
\end{figure}

The impact of the accurate dot product from \cref{eq:fixp-accurate-dotproduct} on ResNet18 is presented in \cref{fig:resnet18-imagenet-fixp-dotproductfunction-top1same}.
Unsurprisingly, better results are achieved using \cref{eq:fixp-accurate-dotproduct} compared to the naive dot product.

\subsection{Summary and hardware impact}

Based on the previous benchmarks and hardware resources presented in \cref{sec:hardware:resources}, we summarize here our results and define the optimum points in terms of number representation, algorithms and hardware resources required.

\begin{figure}[h!]
	\subfloat[ResNet18]{\includegraphics*[width=\columnwidth]{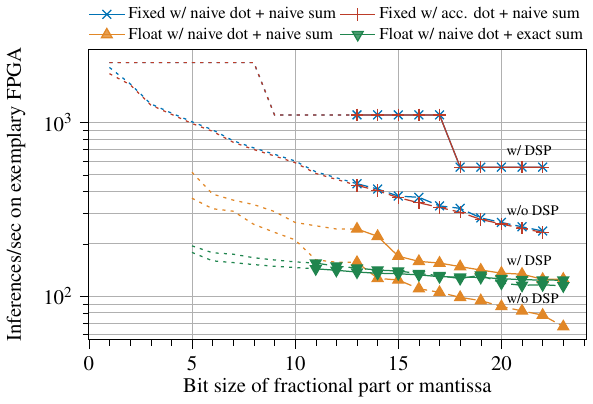}}\\
	\subfloat[MNIST]{\includegraphics*[width=\columnwidth]{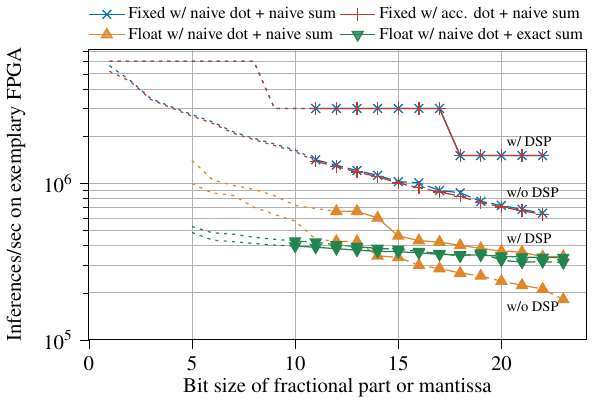}}
	\caption{Hardware performance which can be achieved depending on the number representation, dot product and summation algorithms, with \ac{RNE}. Dashed lines on the left side of the plots represent combinations where the accuracy of the computations is not sufficient to reach a metric of \SI{100}{\percent}.}
	\label{fig:hwsummary}
\end{figure}

\Cref{fig:hwsummary} presents the hardware performance which can be achieved given the different computing parameters evaluated.
From these results, it is clear that using fixed point arithmetic provides the best performance.

\Cref{tab:summary:mnist,tab:summary:resnet18} represent the optimum points for each combination of parameters.
The \emph{"PBits"} column represents the minimum number of bits required for either the mantissa for floating point arithmetic, or the fractional part for fixed point. 
The provided numbers in the \emph{"Estimated inferences/s"} columns are inferred by dividing the available TOPS/TFLOPS at the given size of the fractional part or mantissa by the total MAC operations needed for a single inference on MNIST or ResNet18, respectively.

The dot product according to ORO \cite{Ogita2005} was not implemented in hardware and marked as "\emph{noORO}" in the table. 
Our hardware analysis is based on the AMD IP cores which implement only RNE, hence, we did not analyse RTZ and RNA but we expect very similar results compared to RNE, see "\emph{RNEonly}".
An estimation was done on the implementation of the KN and Pairwise algorithms in Verilog. As these algorithms are more hardware consuming than the implementation of the "\emph{Exact}" sum, we did not further analyze them.

\begin{table}[h!]
\centering
\caption{Summary of the results for MNIST}
\label{tab:summary:mnist}
\begin{booktabs}{@{}l|lllrrr@{}}
	\toprule
	& & & & & \SetCell[c=2]{} {\bf Estimated inferences/s} \\ \cmidrule{6-7}
	& \textbf{Dot} \textbf{Prod.} & \textbf{Sum} & \textbf{Rnd} & \textbf{PBits} & \textbf{w/o DSP} & \textbf{w/ DSP} \\
	\midrule
	\SetCell[r=6]{} \rotatebox[origin=c]{90}{Fixed point} & Accurate & Naive & RNA & 11 & \num{1451925} & \num{3007621} \\
	& Accurate & Naive & RTZ & 11 & \num{1437854} & \num{3007621} \\
	& Naive & Naive & RNA & 11 & \num{1426421} & \num{3007621} \\
	& Naive & Naive & RNE & 11 & \num{1413230} & \num{3007621} \\
	& Accurate & Naive & RNE & 11 & \num{1383330} & \num{3007621} \\
	& Naive & Naive & RTZ & 12 & \num{1352550} & \num{3007621} \\
	\midrule
	\SetCell[r=14]{} \rotatebox[origin=c]{90}{Floating point} & Naive & Naive & RNE & 12 & \num{426014} & \num{662086} \\
	& Naive & Exact & RNE & 10 & \num{428019} & \num{397050} \\ \cmidrule{6-7}
	& ORO \cite{Ogita2005} & Exact & RNE & 10 & noORO & noORO \\
	& ORO \cite{Ogita2005} & KN \cite{Neumaier1974} & RNE & 10 & noORO & noORO \\
	& ORO \cite{Ogita2005} & KN \cite{Neumaier1974} & RNA & 10 & noORO & noORO \\
	& ORO \cite{Ogita2005} & KN \cite{Neumaier1974} & RTZ & 11 & noORO & noORO \\
	& Naive & Exact & RNA & 10 & RNEonly & RNEonly \\
	& Naive & Exact & RTZ & 11 & RNEonly & RNEonly \\
	& Naive & KN \cite{Neumaier1974} & RNE & 10 & noKN & noKN \\
	& Naive & Pairwise & RNE & 10 & noPairwise & noPairwise \\
	& Naive & Pairwise & RNA & 11 & noPairwise & noPairwise \\
	& Naive & Pairwise & RTZ & 11 & noPairwise & noPairwise \\
	& Naive & Naive & RNA & 11 & RNEonly & RNEonly \\
	& Naive & Naive & RTZ & 13 & RNEonly & RNEonly \\
	\bottomrule
\end{booktabs}
\end{table}

\begin{table}[h!]
\centering
\caption{Summary of the results for ResNet18}
\label{tab:summary:resnet18}
\begin{booktabs}{@{}l|lllrrr@{}}
	\toprule
	& & & & & \SetCell[c=2]{} {\bf Estimated inferences/s} \\ \cmidrule{6-7}
	& \textbf{Dot} \textbf{Prod.} & \textbf{Sum} & \textbf{Rnd} & \textbf{PBits} & \textbf{w/o DSP} & \textbf{w/ DSP} \\
	\midrule
	\SetCell[r=5]{} \rotatebox[origin=c]{90}{Fixed point} & Accurate & Naive & RNA & 13 & \num{444} & \num{1106} \\
	& Naive & Naive & RNE & 13 & \num{444} & \num{1106} \\
	& Accurate & Naive & RNE & 13 & \num{434} & \num{1106} \\
	& Accurate & Naive & RTZ & 18 & \num{313} & \num{553} \\
	& Naive & Naive & RTZ & 21 & \num{258} & \num{553} \\
	\midrule
	\SetCell[r=4]{} \rotatebox[origin=c]{90}{Floating p.} & Naive & Naive & RNE & 13 & \num{156} & \num{243} \\
	& Naive & Exact & RNE & 11 & \num{155} & \num{144} \\ \cmidrule{6-7}
	& Naive & KN \cite{Neumaier1974} & RNE & 11 & noKN & noKN \\ 
	& Naive & Pairwise & RNE & 11 & noPairwise & noPairwise \\
	\bottomrule
\end{booktabs}
\end{table}

\acresetall
\section{Conclusion}
\label{sec:conclusion}

We reviewed in this paper various methods for achieving efficient and mathematically robust inference for \acp{NN} in the context of certification of hardware and software for \ac{ML} for aeronautical applications.
This is a challenging task, as special care is required on the mathematical operations to sufficiently accelerate a model in hardware while still preserving the same predictions as the model originally trained on \acp{GPU} or \acp{TPU}.

From a mathematical perspective, we numerically evaluated the various choices which are available, namely: use of floating vs. fixed point arithmetic, reduced precision arithmetic, and more accurate summation and dot product.
From a hardware perspective, we assessed the impact of those choices on the resources required for an exemplary \ac{FPGA}.
Overall, this enabled us to find the good balance in terms of hardware performance and mathematical precision.

Our results show that fixed-point arithmetic with sufficient bits for the fractional part yields the target accuracy for the \ac{NN} and achieves the best performance.

%% file: ms.bbl
\begin{thebibliography}{10}
\providecommand{\url}[1]{#1}
\csname url@samestyle\endcsname
\providecommand{\newblock}{\relax}
\providecommand{\bibinfo}[2]{#2}
\providecommand{\BIBentrySTDinterwordspacing}{\spaceskip=0pt\relax}
\providecommand{\BIBentryALTinterwordstretchfactor}{4}
\providecommand{\BIBentryALTinterwordspacing}{\spaceskip=\fontdimen2\font plus
\BIBentryALTinterwordstretchfactor\fontdimen3\font minus
  \fontdimen4\font\relax}
\providecommand{\BIBforeignlanguage}[2]{{%
\expandafter\ifx\csname l@#1\endcsname\relax
\typeout{** WARNING: IEEEtran.bst: No hyphenation pattern has been}%
\typeout{** loaded for the language `#1'. Using the pattern for}%
\typeout{** the default language instead.}%
\else
\language=\csname l@#1\endcsname
\fi
#2}}
\providecommand{\BIBdecl}{\relax}
\BIBdecl

\bibitem{AIR6988}
``{Artificial Intelligence in Aeronautical Systems: Statement of Concerns},''
  SAE International, Standard SAE ARP AIR6988, Apr. 2021.

\bibitem{ER022}
``{Artificial Intelligence in Aeronautical Safety-Related Systems Statement of
  concerns},'' European Organisation for Civil Aviation Equipment, Standard
  EUROCAE ER-022, May 2021.

\bibitem{EASARoadmap}
``{Artificial Intelligence Roadmap 2.0},'' European Union Aviation Safety
  Agency, Whitepaper, May 2023.

\bibitem{NASARoadmap}
``{Autonomy Verification \& Validation Roadmap and Vision 2045},'' National
  Aeronautics and Space Administration, Technical Memorandum 20230003734, Jan.
  2023.

\bibitem{PerezCerrolaza2023}
J.~Perez-Cerrolaza, J.~Abella, M.~Borg, C.~Donzella, J.~Cerquides, F.~J.
  Cazorla, C.~Englund, M.~Tauber, G.~Nikolakopoulos, and J.~L. Flores,
  ``Artificial intelligence for safety-critical systems in industrial and
  transportation domains: A survey,'' \emph{ACM Comput. Surv.}, Oct. 2023.

\bibitem{CoDANN2020}
EASA and Daedalean, ``Concepts of design assurance for neural networks
  ({CoDANN}),'' Tech. Rep., Mar. 2020.

\bibitem{CoDANN2021}
------, ``Concepts of design assurance for neural networks ({CoDANN}) ii,''
  Tech. Rep., May 2021.

\bibitem{ARP6983}
``{Process Standard for Development and Certification/Approval of Aeronautical
  Safety-Related Products Implementing AI},'' SAE International,
  Work-in-progress standard SAE ARP 6983, Jun. 2023.

\bibitem{Constantinides2002}
G.~Constantinides and G.~Woeginger, ``The complexity of multiple wordlength
  assignment,'' \emph{Applied Mathematics Letters}, vol.~15, no.~2, pp.
  137--140, 2002.

\bibitem{Wang2019}
E.~Wang, J.~J. Davis, R.~Zhao, H.-C. Ng, X.~Niu, W.~Luk, P.~Y.~K. Cheung, and
  G.~A. Constantinides, ``Deep neural network approximation for custom
  hardware: Where we've been, where we're going,'' \emph{ACM Comput. Surv.},
  vol.~52, no.~2, May 2019.

\bibitem{Cherubin2020}
S.~Cherubin and G.~Agosta, ``Tools for reduced precision computation: A
  survey,'' \emph{ACM Comput. Surv.}, vol.~53, no.~2, Apr. 2020.

\bibitem{Kalamkar2019}
D.~Kalamkar, D.~Mudigere, N.~Mellempudi, D.~Das, K.~Banerjee, S.~Avancha, D.~T.
  Vooturi, N.~Jammalamadaka, J.~Huang, H.~Yuen, J.~Yang, J.~Park, A.~Heinecke,
  E.~Georganas, S.~Srinivasan, A.~Kundu, M.~Smelyanskiy, B.~Kaul, and P.~Dubey,
  ``A study of {BFLOAT16} for deep learning training,'' 2019.

\bibitem{Burgess2019}
N.~Burgess, J.~Milanovic, N.~Stephens, K.~Monachopoulos, and D.~Mansell,
  ``Bfloat16 processing for neural networks,'' in \emph{2019 IEEE 26th
  Symposium on Computer Arithmetic (ARITH)}, 2019, pp. 88--91.

\bibitem{Carmichael2019}
Z.~Carmichael, H.~F. Langroudi, C.~Khazanov, J.~Lillie, J.~L. Gustafson, and
  D.~Kudithipudi, ``{Deep Positron}: A deep neural network using the posit
  number system,'' in \emph{2019 Design, Automation \& Test in Europe
  Conference \& Exhibition (DATE)}, 2019, pp. 1421--1426.

\bibitem{Cococcioni2021}
M.~Cococcioni, F.~Rossi, E.~Ruffaldi, S.~Saponara, and B.~Dupont~de Dinechin,
  ``Novel arithmetics in deep neural networks signal processing for autonomous
  driving: Challenges and opportunities,'' \emph{IEEE Signal Processing
  Magazine}, vol.~38, no.~1, pp. 97--110, 2021.

\bibitem{Lu2021}
J.~Lu, C.~Fang, M.~Xu, J.~Lin, and Z.~Wang, ``Evaluations on deep neural
  networks training using posit number system,'' \emph{IEEE Transactions on
  Computers}, vol.~70, no.~2, pp. 174--187, 2021.

\bibitem{Rouhani2020}
B.~Rouhani, D.~Lo, R.~Zhao, M.~Liu, J.~Fowers, K.~Ovtcharov, A.~Vinogradsky,
  S.~Massengill, L.~Yang, R.~Bittner, A.~Forin, H.~Zhu, T.~Na, P.~Patel,
  S.~Che, L.~C. Koppaka, X.~Song, S.~Som, K.~Das, S.~Tiwary, S.~Reinhardt,
  S.~Lanka, E.~Chung, and D.~Burger, ``Pushing the limits of narrow precision
  inferencing at cloud scale with microsoft floating point,'' in
  \emph{Proceedings of the 34th International Conference on Neural Information
  Processing Systems}, ser. NIPS'20, 2020.

\bibitem{Koster2017}
U.~K\"{o}ster, T.~J. Webb, X.~Wang, M.~Nassar, A.~K. Bansal, W.~H. Constable,
  O.~H. Elibol, S.~Gray, S.~Hall, L.~Hornof, A.~Khosrowshahi, C.~Kloss, R.~J.
  Pai, and N.~Rao, ``Flexpoint: An adaptive numerical format for efficient
  training of deep neural networks,'' in \emph{Proceedings of the 31st
  International Conference on Neural Information Processing Systems}, ser.
  NIPS'17, 2017, pp. 1740--1750.

\bibitem{Liu2021}
F.~Liu, W.~Zhao, Z.~He, Y.~Wang, Z.~Wang, C.~Dai, X.~Liang, and L.~Jiang,
  ``Improving neural network efficiency via post-training quantization with
  adaptive floating-point,'' in \emph{Proc. of the IEEE/CVF International
  Conference on Computer Vision (ICCV)}, 2021.

\bibitem{Chen2017}
X.~Chen, X.~Hu, H.~Zhou, and N.~Xu, ``{FxpNet}: Training a deep convolutional
  neural network in fixed-point representation,'' in \emph{2017 International
  Joint Conference on Neural Networks (IJCNN)}, 2017, pp. 2494--2501.

\bibitem{Zhen2019}
Z.~Dong, Z.~Yao, A.~Gholami, M.~Mahoney, and K.~Keutzer, ``{HAWQ}: Hessian
  aware quantization of neural networks with mixed-precision,'' in
  \emph{IEEE/CVF International Conference on Computer Vision (ICCV)}, 2019.

\bibitem{Yao2021}
Z.~Yao, Z.~Dong, Z.~Zheng, A.~Gholami, J.~Yu, E.~Tan, L.~Wang, Q.~Huang,
  Y.~Wang, M.~Mahoney, and K.~Keutzer, ``Hawq-v3: Dyadic neural network
  quantization,'' in \emph{Proceedings of the 38th International Conference on
  Machine Learning}, ser. Proceedings of Machine Learning Research, vol. 139,
  18--24 Jul 2021, pp. 11\,875--11\,886.

\bibitem{Lee2006}
D.-U. Lee, A.~Gaffar, R.~Cheung, O.~Mencer, W.~Luk, and G.~Constantinides,
  ``Accuracy-guaranteed bit-width optimization,'' \emph{IEEE Transactions on
  Computer-Aided Design of Integrated Circuits and Systems}, vol.~25, no.~10,
  pp. 1990--2000, 2006.

\bibitem{Ioualalen2019}
A.~Ioualalen and M.~Martel, ``Neural network precision tuning,'' in
  \emph{Quantitative Evaluation of Systems}, 2019, pp. 129--143.

\bibitem{Martel2017}
M.~Martel, ``Floating-point format inference in mixed-precision,'' in
  \emph{Lecture Notes in Computer Science}, 2017, pp. 230--246.

\bibitem{Benmaghnia2022}
H.~Benmaghnia, M.~Martel, and Y.~Seladji, ``Fixed-point code synthesis for
  neural networks,'' in \emph{Artificial Intelligence, Soft Computing and
  Applications}, Jan. 2022.

\bibitem{Benmaghnia2022b}
------, ``Code generation for neural networks based on fixed-point
  arithmetic,'' \emph{ACM Trans. Embed. Comput. Syst.}, Sep. 2022.

\bibitem{IEEE754}
``{IEEE Standard for Floating-Point Arithmetic},'' \emph{{IEEE Std 754-2019
  (Revision of IEEE 754-2008)}}, pp. 1--84, 2019.

\bibitem{Apostal2020}
S.~F. Jalal~Apostal, D.~Apostal, and R.~Marsh, ``Improving numerical
  reproducibility of scientific software in parallel systems,'' in \emph{IEEE
  International Conference on Electro Information Technology}, 2020.

\bibitem{Robey2011}
R.~W. Robey, J.~M. Robey, and R.~Aulwes, ``In search of numerical consistency
  in parallel programming,'' \emph{Parallel Computing}, vol.~37, no.~4, pp.
  217--229, 2011.

\bibitem{Higham1993}
N.~J. Higham, ``The accuracy of floating point summation,'' \emph{SIAM Journal
  on Scientific Computing}, vol.~14, no.~4, pp. 783--799, 1993.

\bibitem{Kahan1965}
W.~Kahan, ``Further remarks on reducing truncation errors,'' \emph{Commun.
  ACM}, vol.~8, no.~1, p.~40, jan 1965.

\bibitem{Neumaier1974}
A.~Neumaier, ``{Rundungsfehleranalyse einiger Verfahren zur Summation endlicher
  Summen},'' \emph{ZAMM - Journal of Applied Mathematics and Mechanics /
  Zeitschrift für Angewandte Mathematik und Mechanik}, vol.~54, no.~1, pp.
  39--51, 1974.

\bibitem{Dinechin2008}
F.~de~Dinechin, B.~Pasca, O.~Cret, and R.~Tudoran, ``An {FPGA}-specific
  approach to floating-point accumulation and sum-of-products,'' in
  \emph{International Conference on Field-Programmable Technology}, 2008.

\bibitem{Ogita2005}
T.~Ogita, S.~M. Rump, and S.~Oishi, ``Accurate sum and dot product,''
  \emph{SIAM Journal on Scientific Computing}, vol.~26, no.~6, 2005.

\bibitem{Jacob2018}
B.~Jacob, S.~Kligys, B.~Chen, M.~Zhu, M.~Tang, A.~Howard, H.~Adam, and
  D.~Kalenichenko, ``Quantization and training of neural networks for efficient
  integer-arithmetic-only inference,'' in \emph{Proc. of the IEEE Conference on
  Computer Vision and Pattern Recognition (CVPR)}, 2018.

\bibitem{He2016}
K.~He, X.~Zhang, S.~Ren, and J.~Sun, ``Deep residual learning for image
  recognition,'' in \emph{Proc. of IEEE Conference on Computer Vision and
  Pattern Recognition (CVPR)}, 2016, pp. 770--778.

\bibitem{AMD2023}
AMD, ``{UltraScale+ FPGAs Product Selection Guide (XMP103)},'' 2023.

\bibitem{LeCun1998}
Y.~LeCun, L.~Bottou, Y.~Bengio, and P.~Haffner, ``Gradient-based learning
  applied to document recognition,'' \emph{Proceedings of the IEEE}, vol.~86,
  no.~11, pp. 2278--2324, 1998.

\bibitem{Deng2009}
J.~Deng, W.~Dong, R.~Socher, L.-J. Li, K.~Li, and L.~Fei-Fei, ``{ImageNet}: A
  large-scale hierarchical image database,'' in \emph{IEEE Conference on
  Computer Vision and Pattern Recognition}, 2009, pp. 248--255.

\bibitem{IntelNeuralCompressor}
F.~Tian, H.~Chang, H.~Shen, and S.~Chen, ``Intel neural compressor,''
  \url{https://github.com/intel/neural-compressor}, 2022.

\end{thebibliography}
